\documentclass[10pt,twocolumn,letterpaper]{article}

\usepackage[pagenumbers]{cvpr} 

\usepackage{amsmath}
\usepackage{amssymb}
\usepackage{booktabs}
\usepackage{graphicx}
\usepackage{makecell}
\usepackage{multirow}
\usepackage{subcaption}
\usepackage{tabularx}

\usepackage[pagebackref,breaklinks,colorlinks]{hyperref}

\usepackage[capitalize]{cleveref}
\crefname{section}{Sec.}{Secs.}
\Crefname{section}{Section}{Sections}
\Crefname{table}{Table}{Tables}
\crefname{table}{Tab.}{Tabs.}
\newcommand{\specialsize}{\fontsize{8.5}{11}\selectfont}

\def\cvprPaperID{12866} 
\def\confName{CVPR}
\def\confYear{2023}

\usepackage{colortbl}

\begin{document}

\title{Benchmarking and Enhancing Disentanglement in Concept-Residual Models}

\author{Renos Zabounidis$^{1}$, Ini Oguntola$^{1}$, Konghao Zhao$^{2}$, Joseph Campbell$^{1}$, Simon Stepputtis$^{1}$, Katia Sycara$^{1}$\\
Carnegie Mellon University$^{1}$,Wake Forest University$^{2}$\\
{\tt\small \{renosz, ioguntol, jcampbell, sstepput, katia\}@cs.cmu.edu}$^{1}$, {\tt\small zhaok220@wfu.edu }$^{2}$\\
}
\maketitle

\begin{abstract}
Concept bottleneck models (CBMs) are interpretable models that first predict a set of semantically meaningful features, i.e., concepts, from observations that are subsequently used to condition a downstream task. However, the model's performance strongly depends on the engineered features and can severely suffer from incomplete sets of concepts. Prior works have proposed a side channel -- a residual -- that allows for unconstrained information flow to the downstream task, thus improving model performance but simultaneously introducing information leakage, which is undesirable for interpretability. This work proposes three novel approaches to mitigate information leakage by disentangling concepts and residuals, investigating the critical balance between model performance and interpretability. Through extensive empirical analysis on the CUB, OAI, and CIFAR 100 datasets, we assess the performance of each disentanglement method and provide insights into when they work best. Further, we show how each method impacts the ability to intervene over the concepts and their subsequent impact on task performance.
\end{abstract}

\section{Introduction}

State-of-the-art deep learning models have achieved remarkable performance in complex tasks, surpassing humans in areas such as computer vision~\cite{han2022survey}, speech recognition~\cite{radford2023robust}, and strategic game playing~\cite{alphago, brown2018superhuman}.
However, despite their high accuracy, the underlying decision-making process of these models is not readily understandable to humans, and as such, they are often treated as ``black boxes''.
This lack of interpretability introduces risks when applying deep learning models to safety-critical tasks such as medical imaging~\cite{salahuddin2022transparency}, as it is challenging to understand how a model arrived at a prediction or when it might fail.

Concept Bottleneck Models (CBMs)~\cite{koh2020concept} inject interpretability into deep neural networks by first predicting a set of semantically meaningful concepts, and then using them for subsequent prediction.
For these models to be accurate, they must learn to recognize a sufficiently comprehensive set of concepts for the task at hand.
This is a strong constraint in practice, as defining concepts and acquiring a sufficient amount of data to train them is challenging.
To mitigate this issue, prior methods have proposed the usage of an additional side-channel of residual information~\cite{yuksekgonul2022post, sawada2022concept, zabounidis2023concept, oguntola2023theory, zhang2023decoupling} in downstream classification, which we refer to as a Concept-Residual Bottleneck Model (CRBM).
This is intended to obviate the need for a fully comprehensive concept set as the residual layers learn to encode any necessary, but missing information.

However, residuals are susceptible to \textit{information leakage} in which the residual layer can re-encode information already contained within the set of concepts.
This reduces the interpretability of CRBMs as the downstream classification layers may use the information in the residual rather than the concepts, thus limiting our ability to understand what information the network is reasoning with.

\begin{figure}
    \centering
    \includegraphics[width=1\linewidth]{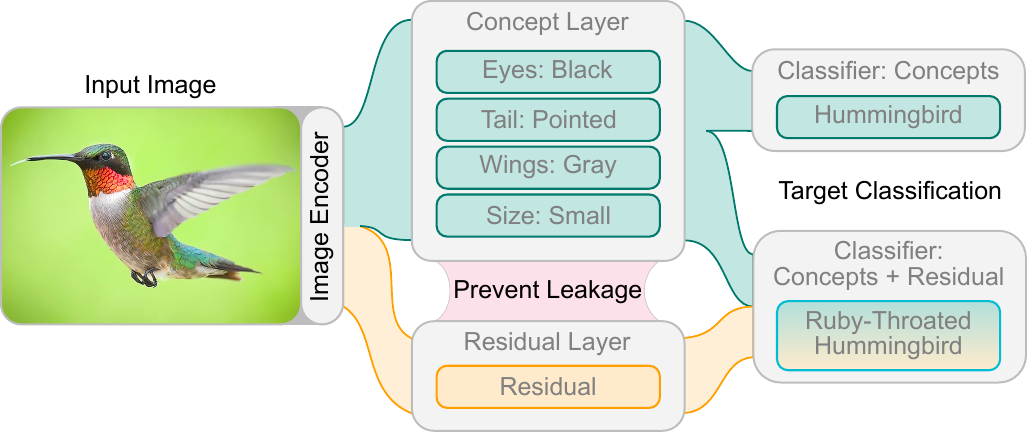}
    \caption{
    Our inference pipeline consists of two main components: the Concept layer, encoding primary image features, and the Residual layer, capturing additional information. We compare various methods to mitigate information leakage between these components. The additional information encoded in the residual leads to improved classification accuracy. }
    \label{fig:method_overview}
\end{figure}

In this work, we propose a systematic and comprehensive analysis of concept-residual information leakage which, to the best of our knowledge, is the first of its kind.
We propose a set of proxy metrics to measure interpretability based on concept interventions -- where an external oracle edits the predicted concepts and propagates the changes to the downstream classifier.
Using these metrics, we investigate three methods for disentangling the concept and residual layers in order to prevent loss of interpretability when a residual side-channel is used, as shown in Fig.~\ref{fig:method_overview}.

Concretely, our contributions are as follows:

\begin{enumerate}
    \item We systematically analyze four variants of CRBMs over four widely-used datasets to understand how we can best disentangle concept and residual layers to reduce information leakage.
    \item We study CRBMs in two types of tasks -- those with a complete set of concepts and those with an incomplete set of concepts -- to better understand performance characteristics.
    \item We propose the use of concept and residual interventions that act as a proxy measurement of the interpretability of a CBM.
\end{enumerate}

\section{Related work}

The utilization of human concepts as interpretative tools in neural networks is a growing field, with research focusing on how these networks encode and utilize concepts \cite{lucieri2020interpretability, kim2018interpretability, mcgrath2021acquisition}. Mao et al. learned novel visual concepts and their interaction with other concepts in a few-shot manner from images with sentence descriptions \cite{mao2015learning}. Ghandeharioun et al. generate concept traversals (CTs), defined as a sequence of generated examples with increasing degrees of concepts that influence a classifier’s decision \cite{ghandeharioun2021dissect}.
Han et al. propose the visual concept-metaconcept learner (VCML) for joint learning of concepts and metaconcepts from images and associated question-answer pairs \cite{han2019visual}. Higgins et al. create a variational autoencoder (VAE) framework capable of learning disentangled factors (i.e. concepts) \cite{higgins2016early}.

Originating from concepts-first methodologies \cite{lampert2009learning, kumar2009attribute}, Concept Bottleneck Models (CBMs), like those described by Koh et al. \cite{koh2020concept}, predict downstream tasks using an intermediary bottleneck layer trained with supervised losses. Baradori et al. extended CBMs to denoise the concepts \cite{bahadori2020debiasing}. ProbCBM models uncertainty in concept prediction and provides explanations based on them and their corresponding uncertainty \cite{kim2023probabilistic}. Shin et al. develop various strategies for selecting which concepts maximally improve intervention effectiveness \cite{shin2023closer}. Chauhan et al. similarly develop a method combining concept prediction uncertainty and influence of the concept on the final prediction to improve intervention effectiveness \cite{chauhan2023interactive}.
CBMs offer interpretability advantages but are limited by the need for concept labels during training. The challenge lies in creating a fully representative set of concepts for a domain. When concepts are not expressive enough for the task, performance suffers.

Studies have shown that CBMs often encode information beyond the intended concepts \cite{mahinpei2021promises, margeloiu2021concept}. While Chen et al. attempted to align concepts with individual dimensions with IterNorm \cite{chen2020concept}, it is not clear to what extent this approach helped with achieving disentanglement.

The integration of residual layers in CBMs has been explored to ease the requirement that training concept sets are complete for a task. Yuksekgonul et al. add a non-decorrelated residual predictor to their model, complementing transferred concepts \cite{yuksekgonul2022post}. Omitting this predictor enables assessment of the model's fundamental interpretability, though no decorrelation is applied otherwise. Sawada et al. proposed a residual predictor, but did not directly address leakage \cite{sawada2022concept}. Zabounidis et al. \cite{zabounidis2023concept} applied CBMs in multi-agent reinforcement learning, noting improved intervention performance with disentanglement techniques. Oguntola et al. utilized mutual information to reduce intra-concept leakage in reinforcement learning agents, reporting performance improvements in task reward \cite{oguntola2023theory}. Similarly, DCBMs \cite{zhang2023decoupling} proposed algorithms based on mutual information for automatic label correction and concept tracing but did not focus explicitly on leakage issues. These methods hint at the need for minimizing concept-residual leakage, and some even propose solutions. However, to our knowledge no paper has evaluated the actual impact of decorrelation methods on concept-residual information leakage. Similarly, these works do not investigate the relationship between information leakage and interventions. Out work aims to fill this gap, providing a detailed analysis and quantification of decorrelation methods' impact on reducing concept-residual leakage and exploring the interplay between information leakage and intervention effectiveness, thus contributing significantly to our understanding of concept-residual leakage and practical application in the domain.

\section{Preliminaries}
\paragraph{Concept Bottleneck Models}
Concept bottleneck models (CBMs) are designed for predicting a target variable \( y \in \mathbb{R} \) from an input \( x \in \mathbb{R}^d \). These models operate by mapping the input \( x \) to a concept space \( c \in \mathbb{R}^k \), with \( k \) concepts, and using these concepts to predict \( y \). Formally, a CBM is a composite function \( f(g(x)) \), where \( g: \mathbb{R}^d \to \mathbb{R}^k \) maps \( x \) to the concept space, and \( f: \mathbb{R}^k \to \mathbb{R} \) maps the concepts to the final prediction.

In CBMs, two types of accuracies are considered: task accuracy, the accuracy of \( f(g(x)) \) in predicting \( y \), and concept accuracy, the accuracy of \( g(x) \) in predicting the concepts \( c \). We focus on an \textit{Independent} Bottleneck: This involves training \( \hat{f} \) and \( \hat{g} \) separately. $( \hat{f} )$ is trained using the true concepts $( c ),$ and $( \hat{g} )$ is trained to predict the concepts from \( x \). Training then includes a hyperparameter \( \lambda \) which balances the focus between concept and task accuracy. 

CBMs can be implemented by modifying an existing neural network to include a layer with \( k \) neurons (matching the number of concepts) and training this network using one of the above methods. The overall loss function for a CBM can be represented as:
\begin{align}
 \mathcal{L}_{\text{cbm}} = \sum_i^N \Bigl( \mathcal{L}_y(f(x_i),y_i) +  \lambda \sum_j^{D_{\text{ex}}} \mathcal{L}_j(c_j^{\text{ex}}(x_i), c_{i,j}) \Bigr)   
\end{align}
Here, \( \mathcal{L}_y \) is the loss function for the target prediction, \( \mathcal{L}_j \) is the loss function for the \( j \)-th concept prediction, and \( \lambda \) is the hyperparameter balancing the two types of losses.

\paragraph{Concept-Residual Bottleneck Models}
To generalize bottleneck models, CRBMs introduce a residual layer. This addition enables concept models to effectively capture more complex data patterns while still retaining the essence of CBMs. Standard CBMs can be considered as a special case with residual size set to zero.

In concept models, the input \( x \) is mapped to two distinct spaces: the concept space \( c \) and the residual space \( r \). The formulation of a concept model is given by:
\begin{align}
\hat{y} = f(g(x), r(x))
\end{align}
Here, \( g: \mathbb{R}^d \to \mathbb{R}^k \) maps \( x \) to the concept space, while \( r: \mathbb{R}^d \to \mathbb{R}^m \) captures residual information not represented in \( c \). The function \( f: \mathbb{R}^k \times \mathbb{R}^m \to \mathbb{R} \) combines these representations for the final prediction.

\section{Methodology}
Our work focuses on decorrelating and benchmarking performance on Concept-Residual Bottleneck Models. With the addition of a residual layer, traditional CBM training approaches like sequential training become impractical; to this end, we use what we call \textit{semi-independent training}.

Under this training paradigm, gradients are not allowed to backpropagate through the concept layer, only through the residual layer. During training, the target network receives ground truth concepts, not the estimated ones. The other possible training paradigm would be joint training, where predicted concepts are passed through to the downstream task. We do not use this paradigm as it would introduce the issue of intra-concept information leakage \cite{mahinpei2021promises}. Our methodology therefore allows us to focus soley on concept-residual information leakage.

\textit{Concept-residual leakage} refers to the scenario where information from the concept layer seeps into the residual layer or vice versa, which can compromise the model's interpretability and performance. To address concept-residual leakage, we study the effects of different disentanglement techniques, aiming to maintain statistical independence between concepts and residual. These techniques, detailed in subsequent subsections, include:
\begin{itemize}
\item \textbf{Regularization Methods}: Introducing regularization terms in the loss function to penalize dependencies between different concepts.
\item \textbf{Architectural Strategies}: Designing the network architecture to directly separate concept representations, reducing their likelihood of intermingling.
\end{itemize}

\paragraph{Iterative normalization}
We will begin by discussing Iterative Normalization, an architectural approach that utilizes whitening as a means of disentanglement. To mitigate concept-residual leakage, prior work focuses on decorrelating the neuron activation vectors via whitening, particularly within the residual layer \( r(\cdot) \) \cite{zabounidis2023concept}. This is achieved through \textit{iterative normalization}, as proposed by Huang et al. \cite{huang2019iterative}. Given a matrix \(\mathbf{X} \in \mathbb{R}^{b \times (j+k)}\) consisting of the concatenated concept and residual vectors over a mini-batch of \( b \) samples, we produce a whitened matrix \(\mathbf{X}'\) using Zero-Phase Component Analysis (ZCA) whitening:
\begin{equation}
    \mathbf{X}' = \mathbf{D} \Lambda^{-\frac{1}{2}} \mathbf{D}^T (\mathbf{X} - \mu_{\mathbf{X}})
\end{equation}
Here, \(\mathbf{D}\) and \(\Lambda\) represent the eigenvectors and eigenvalues of the covariance matrix of \(\mathbf{X}\), respectively. The iterative normalization technique employs an iterative optimization approach to progressively whiten \(\mathbf{X}\). The hyperparameter \( T \) controls the number of optimization iterations, allowing for partial decorrelation by setting \( T \) to a lower value, such as \( T = 2 \). This flexibility is crucial for stabilizing the training process, as a higher \( T \) can introduce excessive normalization disturbance, reducing training efficiency, a phenomenon particularly pronounced in Multi-Agent Reinforcement Learning (MARL) contexts \cite{huang2019iterative}. Our approach decorrelates the concept and residual layers effectively, integrating the whitening process into each training iteration, thus obviating the need for additional optimization steps as required in previous methodologies \cite{chen2020concept}.

\paragraph{Cross-correlation minimization}
To minimize the correlation between concept and residual representations, we introduce a decorrelation loss, formulated to enforce orthogonality between these components. Unlike Iter Norm, \cite{huang2019iterative} this loss does not also enforce intra concept decorrelation and is a regularization method, rather than architectural. The loss function is derived from the covariance matrix of the concept and residual outputs, \( C \) and \( R \), over a mini-batch of \( b \) samples. This matrix is defined as:
\begin{equation}
    \mathbf{Cov}_{CR} = \frac{1}{b} C^T R
\end{equation}

Our decorrelation loss, \(\mathcal{L}_{decorr}\), penalizes the off-diagonal elements of \(\mathbf{Cov}_{CR}\), ensuring that the concept and residual layers capture distinct and non-overlapping information. The loss function is defined as the sum of the squared off-diagonal elements of \(\mathbf{Cov}_{CR}\):
\begin{equation}
    \mathcal{L}_{decorr} = \sum_{i \neq j} (\mathbf{Cov}_{CR})_{ij}^2
\end{equation}
This approach ensures that learned representations \( C \) and \( R \) are orthogonal, thus reducing information overlap and enhancing the distinctiveness of each layer.

\paragraph{Minimizing mutual information with CLUB}
Both iterative normalization and cross correlation minimize a correlation -- a linear metric. This means both methods potentially fail to capture non-linear dependencies, and thus intra-concept leakage can still occur. We therefore turn to mutual information (MI) as a more robust measure of statistical dependence. MI is uniquely advantageous in that it is zero \textit{if and only if} the random variables in question are independent. This characteristic makes MI an ideal candidate for ensuring the independence of concept and residual representations in our model. The mathematical representation of MI as a form of KL-divergence is as follows:
\begin{align}
    I(C;R) = D_{KL}(\mathbb{P}_{CR} \parallel \mathbb{P}_C \otimes \mathbb{P}_R)
\end{align}

To effectively minimize MI, we employ a contrastive log-ratio upper bound method \cite{oguntola2023theory}, originally proposed in Cheng et al. \cite{cheng2020club}. This approach utilizes variational inference to approximate the conditional distribution between concept (\(\mathbf{c}\)) and residual (\(\mathbf{r}\)) vectors, which we can then use to estimate the mutual information. The variational loss functions are formulated with policy parameters \( \sigma \) and variational parameters \( \theta \), as follows:
\begin{align}
    \mathcal{L}_{variational}(\theta) &= -\mathbb{E}_{p_\sigma(\mathbf{c}, \mathbf{r})}[\log q_\theta(\mathbf{r} \mid \mathbf{c})] \\
    \mathcal{L}_{MI}(\sigma, \theta) &= \mathbb{E}_{p_\sigma(\mathbf{c}, \mathbf{r})}[\log q_\theta(\mathbf{r} \mid \mathbf{c})] \nonumber \\
    &\quad -  \mathbb{E}_{p_\sigma(\mathbf{c})} \mathbb{E}_{p_\sigma(\mathbf{r})}[\log q_\theta(\mathbf{r} \mid \mathbf{c})]
\end{align}
In these equations, \( p_\sigma \) denotes the distribution of concepts and residuals as influenced by the current model parameters, and \( q_\theta(\mathbf{r} \mid \mathbf{c}) \) represents a variational approximation of the true conditional distribution. This approach allows for a controlled reduction of MI between concept and residual layers, thereby promoting their independence and enhancing the interpretability of the overall model \cite{oguntola2023theory}.

\section{Quantifying concept-residual leakage}
After outlining several strategies aimed at mitigating concept-residual leakage, we move to establish precise methodologies for its evaluation.
Our initial challenge is selecting appropriate datasets. A key issue in the development of CRBMs (Concept-Residual Bottleneck Models) is adapting and creating datasets with an incomplete set of concepts. Traditional concept bottleneck model datasets feature \textit{Complete Concept Sets} \cite{oai, WahCUB_200_2011, koh2020concept}. These sets are sufficiently representative to be considered complete in terms of final performance. When training a CRBM, the residual will either encode no additional information, or it will be affected by concept-residual leakage. Consequently, it becomes challenging to discern whether techniques for reducing concept-residual leakage inadvertently diminish the expressiveness of the residual layer. Therefore, it is essential to employ datasets with \textbf{Incomplete} Concept Sets.

\paragraph{Incomplete Concept Sets} In defining incomplete concept sets, it is crucial to differentiate between \textbf{performant} and \textbf{nonperformant} concept sets. Performant concept sets include useful concepts for the purposes of the downstream task. The performance of a concept set can be gauged by comparing the accuracy of a bottleneck model, trained on these concepts, with the expected random accuracy for the downstream task. If the model's performance is comparable to random, the concepts are deemed nonperformant. We will later present an example of such a dataset in our results. In contrast, performant concept sets, although incomplete, are applicable for the downstream task. An illustration of this is our adaptation of the CIFAR 100 dataset, where the concepts represent one of 20 superclasses. Knowing a superclass narrows the final class choice to one of five options, providing a 20\% accuracy. To achieve 100\% accuracy, an additional concept is needed to define the conditional distribution of the remaining five classes given the superclass identity. However, understanding the superclass objectively offers more information than a random selection, which in the case of CIFAR 100, would be only 1\%.

\subsection{Metrics for concept-residual leakage}
Quantifying the information leakage between the concept layer and the residual layer remains a formidable challenge in concept model research. While metrics such as cross-correlation minimization and mutual information (MI) minimization have been proposed to measure this leakage, these methods provide limited insights \cite{mahinpei2021promises}. Cross-correlation focuses on linear dependencies, and MI extends this to non-linear relationships. However, neither metric, despite their practicality, might completely represent the intricate dynamics of information leakage in real-world applications. Moreover, it remains uncertain what threshold values for these metrics are necessary to effectively prevent concept-residual leakage in a given dataset. We argue that while these metrics are beneficial for quantifying information leakage, a more reliable measure of a model’s susceptibility to concept-residual leakage is evaluated through its performance in three types of interventions.

\paragraph{Positive concept interventions.}
In positive concept interventions, we replace the concept layer with accurate, true concepts. The impact of this varies depending on the concept set. For complete sets, positive interventions are expected to yield similar performance to bottleneck models without information leakage, since all necessary information for the downstream task is already encoded. In this scenario, the best outcome for CRBMs is to match the performance of CBMs, with any performance gap indicating leakage. For incomplete but performant sets, CRBMs should outperform CBMs during interventions. While models with minimal intra-concept leakage are hypothesized to perform better, there isn't a clear benchmark for leakage-free performance. Therefore, evaluation of other intervention methods is crucial. In contrast, with nonperformant, incomplete sets, decorrelation techniques are likely to degrade performance, as ineffective concepts restrict representational capabilities of the residual when decorrelated.

\paragraph{Random concept interventions.}
Random concept interventions involve replacing the concept predictions for a given input with those from another randomly chosen, in-distribution input. When dealing with complete concept sets, CRBMs that effectively utilize concepts and have minimal intra-concept leakage should exhibit a significant decline in performance under these conditions. Specifically, for complete sets in CRBMs without concept-residual leakage, intervention performance is anticipated to fall to near-random levels. With incomplete sets, the extent of performance decline hinges on the nature of the residual information. If the concept and residual are independent, the drop should align with the concept set's contribution. However, if the residual represents a conditional distribution based on the concept set, the performance might plummet to nearly random levels. In both scenarios, a decrease in performance is expected. Conversely, for nonperformant, incomplete sets, we expect negligible performance change, as the concepts should not be used in the model's decision process.

\paragraph{Random residual interventions.}
In random residual injection interventions, we introduce arbitrary residual information into the model. With complete concept sets, models exhibiting minimal inter concept-residual leakage are expected to retain their performance, as they primarily rely on the complete concept set. However, for incomplete, performant concept sets, such models may undergo a decrease in performance. For unperformant, incomplete concept sets, we expect accuracy to go to near random, as the residual is the only part of the bottleneck encoding useful information.

\section{Experiments}

We evaluate Concept Models using different decorrelation techniques in challenging image classification and medical settings, demonstrating how Concept Models maintain performance while improving intervention performance and interpretability. We used the following datasets to evaluate Concept Models systematically:

\begin{table*}[ht]
    \centering
    \specialsize
    \setlength{\tabcolsep}{0.65em}%
    \begin{tabular}{c | 
        cccc 
        >{\columncolor[gray]{0.95}}c>{\columncolor[gray]{0.95}}c>{\columncolor[gray]{0.95}}c>{\columncolor[gray]{0.95}}c |
        cccc 
        >{\columncolor[gray]{0.95}}c>{\columncolor[gray]{0.95}}c>{\columncolor[gray]{0.95}}c>{\columncolor[gray]{0.95}}c 
        }
        \toprule
        & \multicolumn{8}{c|}{Incomplete Concept Sets} & \multicolumn{8}{c}{Complete Concept Sets} \\ 

            & \multicolumn{4}{c}{\makecell{CIFAR 100}} & \multicolumn{4}{>{\columncolor[gray]{0.95}}c|}{CUB - No Majority Voting} & \multicolumn{4}{c}{CUB} & \multicolumn{4}{>{\columncolor[gray]{0.95}}c}{OAI} \\ 

        Method & B & C$^{+}\!\uparrow$ & C$^{-}\!\downarrow$ & R$^{-}\!\uparrow$ & B & C$^{+}\!\uparrow$ & C$^{-}\!\downarrow$ & R$^{-}\!\uparrow$ & B & C$^{+}\!\uparrow$ & C$^{-}\!\downarrow$ & R$^{-}\!\uparrow$ & B & C$^{+}\!\uparrow$ & C$^{-}\!\downarrow$ & R$^{-}\!\uparrow$ \\ 
        \midrule
        Bottleneck              & 0.11 & 0.20 & 0.01 & - & 0.47 & 0.37 & 0.02 & - & 0.75 & 0.99 & 0.02 & - & 0.69 & 0.96 & 0.39 & - \\ \hline
        Latent                  & 0.60 & 0.72 & 0.59 & 0.02 & 0.76 & 0.77 & 0.74 & 0.02 & 0.79 & 0.93 & 0.75 & 0.04 & 0.72 & 0.78 & 0.67 & 0.46 \\
        Decorr.            & 0.60 & 0.73 & 0.59 & 0.02 & 0.76 & 0.77 & 0.72 & 0.02 & 0.79 & 0.94 & 0.61 & 0.16 & 0.72 & 0.82 & 0.56 & 0.57 \\
        IterNorm               & 0.60 & 0.68 & 0.59 & 0.02 & 0.77 & \textbf{0.79} & 0.74 & 0.02 & 0.79 & \textbf{0.96} & 0.60 & 0.22 & 0.71 & 0.90 & \textbf{0.38} & 0.56 \\
        MI             & 0.60 & \textbf{0.83} & \textbf{0.08} & \textbf{0.11} & 0.73 & 0.75 & \textbf{0.54} & \textbf{0.05} & 0.76 & 0.96 & \textbf{0.19} & \textbf{0.44} & 0.67 & \textbf{0.94} & 0.40 & \textbf{0.66} \\
        \bottomrule
    \end{tabular}
    \caption{
    Classification accuracy with different disentanglement methods given positive interventions (C$^+$), negative interventions (C$^-$), and randomized residuals (R$^-$). Experiments were run over three seeds, reporting average results.
    }
    \label{tab:performance}
\end{table*}

The \textbf{CIFAR-100} dataset is uniquely utilized by employing its 20 superclasses as primary concepts \cite{krizhevsky2009cifar}. Given a superclass, an image can belong to any of the five associated classes, indicating that knowledge of the superclass alone is insufficient for precise classification, as the concept space is incomplete. The challenge lies in predicting one of 100 classes, which requires the residual to encode the conditional distribution of possible classes given a superclass. This setup makes CIFAR-100 an ideal benchmark for evaluating methods that can disentangle the residual and concept layers while preserving valuable information in both.

The \textbf{CUB} (Caltech-UCSD Birds) dataset is a rich resource for fine-grained visual categorization, particularly in bird species recognition \cite{welinder2010caltech}. It contains images of 200 bird species, each annotated with detailed attributes like feather color, beak shape, and wing patterns. These concepts are learned via multi-binary classification. Bottleneck Models trained on CUB rely heavily on the use of majority voting \cite{chen2020concept}. Chen et al. explain that the provided concepts are noisy, and thus to make them effective for CBMs they employ majority voting; if more than 50\% of a downstream class has a particular concept in the data, then we set all that downstream task to have that concept. In order to investigate what occurs when CRBMs are trained on noisy incomplete concepts, we train an additional variant of CUB -- \textit{CUB (No Majority Voting)} -- using the original concept labels for each sample.

The \textbf{OAI (Osteoarthritis Initiative)} dataset, comprising 36,369 knee X-ray data points, focuses on individuals at risk of knee osteoarthritis. It includes both radiological and clinical data, with the primary task being to predict the Kellgren-Lawrence grade (KLG), a four-level ordinal scale used by radiologists to assess osteoarthritis severity. Higher KLG scores indicate more severe disease. The dataset utilizes ten ordinal variables as concepts for analysis, encompassing joint space narrowing, bone spurs, calcification, and other clinical features. These variables, essential for evaluating osteoarthritis severity, align with the preprocessing techniques employed by Pierson et al. \cite{pierson2021algorithmic}.

\paragraph{Models}
We use a convolutional network pre-trained on ImageNet as the backbone for the feature extractor. The output of this layer is split which, at the last layer, splits into the concept and residual layers. The target network is one layer for CIFAR and CUB, and three for OAI. Further architecture details can be found in the Appendix. 

\paragraph{Experimental design}

We train CRBMs to test under which methods perform the best with minimal concept-residual leakage. We train models of varying residual size from 0 (bottleneck) up to 64 in powers of 2. We use these trained models to apply our testing methodology for evaluating the concept-residual information leakage by performing positive interventions, negative concept interventions, and negative residual interventions. We compare these results with traditional dependence metrics of cross correlation and mutual information. Mutual information is approximated post hoc for all models using the CLUB estimator \cite{cheng2020club}.

\subsection{Results}

\begin{figure}[tb]
\includegraphics[width=0.46\textwidth]{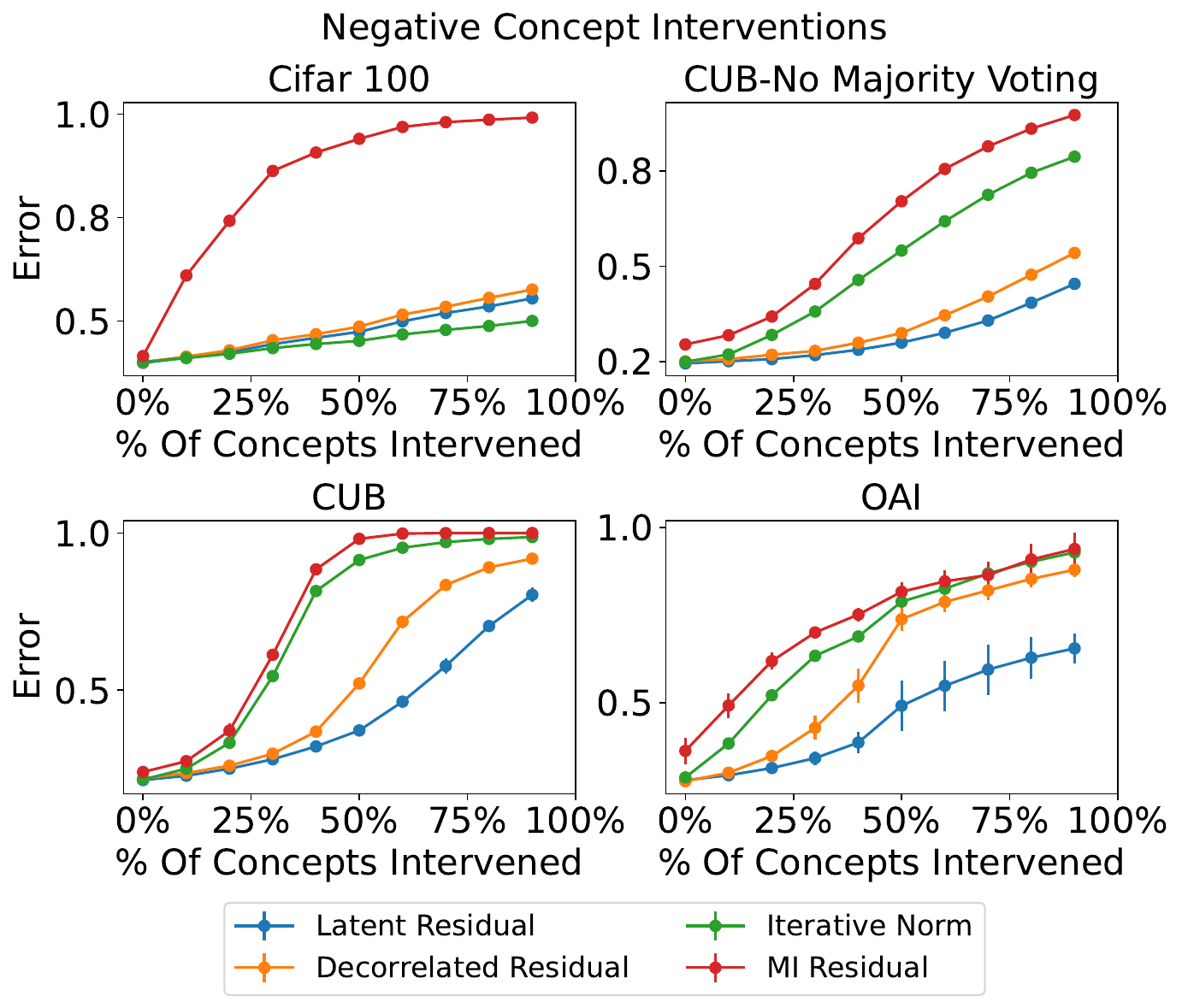}
\vspace{-2mm}
\caption{Classification error with negative interventions (replacing predicted concepts with incorrect values).}
\label{fig:neg_interventions}
\end{figure}

\begin{figure*}[tb]
  \centering
  \includegraphics[width=\textwidth]{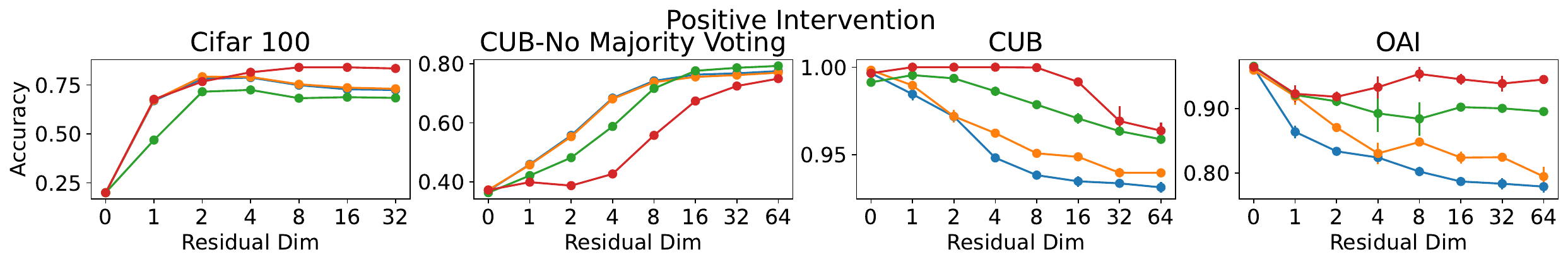}\\
  \vspace{-1mm}
  \includegraphics[width=\textwidth]{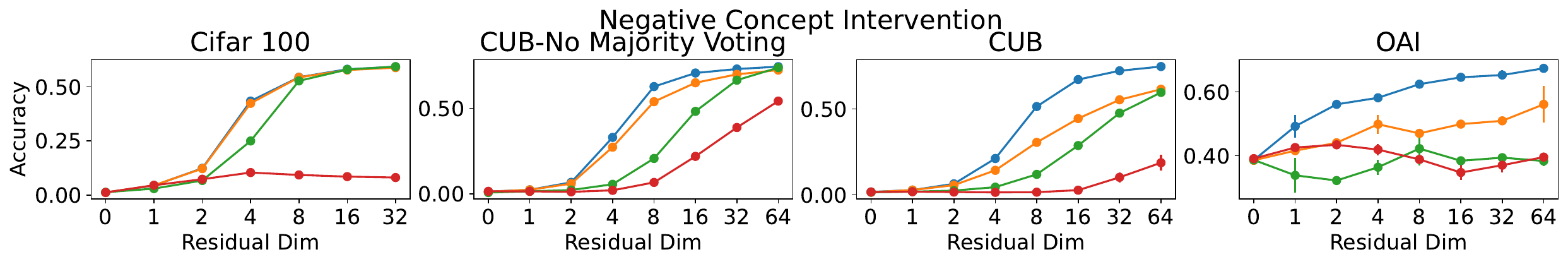}\\
  \vspace{-1mm}
  \includegraphics[width=\textwidth]{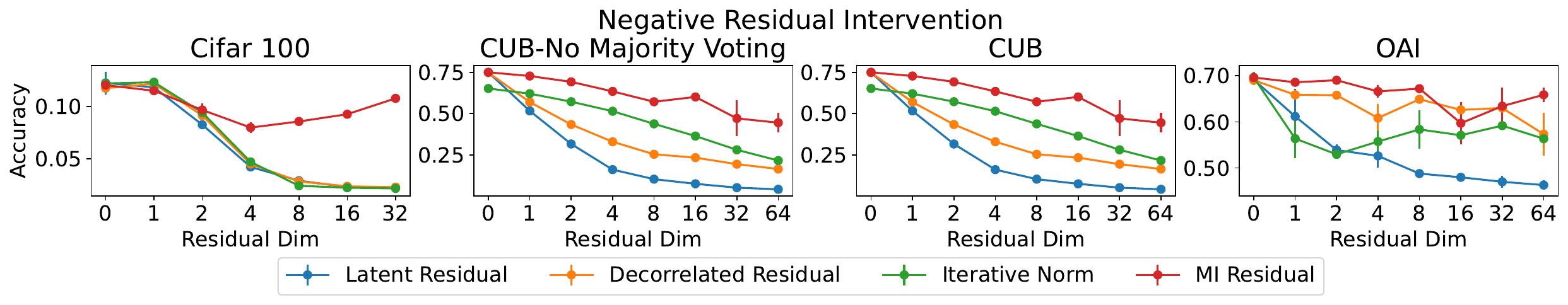}\\
  \vspace{-2mm}
  \label{fig:positive}
  \caption{Test-time intervention results are averaged over 3 seeds, with standard deviation shown in bar plots. \textbf{Top}: Positive interventions assess information leakage; a drop in performance indicates reduced reliance on concepts. \textbf{Middle}: Negative Concept interventions; lower performance reflects higher concept dependence, with MI Residual effectively minimizing information leakage. \textbf{Bottom}: Negative residual interventions; lower performance or alignment with baseline suggests reduced concept-residual leakage and lower information leakage.}
  \label{fig:overall}
\end{figure*}

\paragraph{CRBMs for complete concept sets.}
When concept sets are complete, the residual is not expected to be able to encode any additional useful information. We thus expect that increasing residual size will only degrade performance, and that the extent of this degradation is based on the models concept-residual leakage. In Table \ref{tab:performance}, we show that for datasets where concepts are complete, the best CRBM decorrelation methods meet baseline performance on every dataset, with results across the methods generally performing the same. We hypothesize this is due to the fact that without interventions, CRBMs have the flexibility of relying on a mix of representations from the concept layer and the residual layer. Interpretability is still decreased if information leakage between the two exists, but performance does not degrade. Positive interventions are where we see the negative effects of concept-residual information leakage. If the residual doesn't encode any concept-related information, and the concepts are important to the task, then interventions will affect the downstream process. Perfect disentanglement would therefore be an upper bound on performance because the downstream model \textit{has} to use the concepts, while complete replication of concepts would be a lower bound, as there is nothing forcing the downstream model to use the concepts instead of the residual. In Table \ref{tab:performance}, we see that the latent residual method has substantially reduced intervention performance compared to decorrelation techniques such as IterNorm and Mutual Information. These two methods still do not quite match the bottleneck model for the largest residual size we report, but they come within a couple percent.

\paragraph{Selecting residual size for complete concept sets.}
CRBM performance decreases with increasing residual dimension. In both CUB and OAI Figure \ref{fig:overall} shows that methods like MI and IterNorm are able to maintain performance, while methods like Decorrelation and Latent Residual decrease in performance steadily. This suggests that MI and IterNorm are better minimizers of concept-residual leakage. For practical consideration, CRBMs do not improve intervention performance, but through decorrelation techniques can still perform comparably to bottleneck models.

\paragraph{CRBMs for incomplete concept sets.} There are two main categories of incomplete concept sets: those with performant concepts, and those with concepts that are too noisy for the downstream model to effectively utilize. For both types of incomplete concept sets, we expect to see improved baseline performance over the bottleneck, as the residual layer is able to encode information missing in the concept set. We see this clearly in both CIFAR-100 and CUB (without majority voting), where baseline performance of CRBMs increase as much as 6x over their bottleneck counterparts. For performant concept sets, we expect that interventions will still improve performance further, as the residual layer is only supplementing representations from the concept layer. CIFAR-100 is an example of a dataset with performant concepts, and we see that performance again increases strongly from baseline to positive interventions, especially for the MI method, which we later hypothesize best minimizes mutual information. CUB with no majority voting is an example of concepts which are not performant. We find that for CRBMs, intervention performance is negligibly higher then the bottleneck. Notably, the bottleneck performs even worse on the task then baseline. We hypothesize that in this instance, CRBMs learn to ignore the concept model and rely solely on the residual layer. This conclusion is further supported when we look at negative interventions.

\paragraph{Selecting residual size for incomplete concept sets.}
Residual size is an important consideration for CRBMs in incomplete concept sets. In this setting, the residual layer must be large enough to encode the additional representations needed by the downstream policy. At the same time, increasing the residual layer increases the potential information leakage. This underscores the necessity for decorrelation techniques which maintain low levels of concept-residual leakage across increasingly large residual sizes. For incomplete concept sets with performant concepts like CIFAR-100, we see lower performing techniques like latent residual and decorrelated residual peak in performance at around a residual dimension of around 2 and then degrade. MI is able to steadily increase intervention performance at increasing residual dimensions, underscoring the importance of disentanglement techniques.

For incomplete concept sets with non-performant concepts, every method steadily increases in performance as residual size increases. This is because the concept layer is negligibly used by the downstream model and therefore concept leakage has no effect on performance. As we show with negative interventions (Figure \ref{fig:neg_interventions}), only MI performs notably different when random concepts are injected. In this case, disentanglement only hurts performance, forcing the downstream model to use incorrect concepts. 

\paragraph{Positive interventions are not enough.}
While positive interventions do improve when we minimize information leakage, models with concept-residual leakage still perform above the baseline. Table \ref{fig:overall} shows that across all three datasets with performant concepts (CIFAR 100, CUB, and OAI), the latent residual is still able to perform positive interventions much higher then baseline performance. Having positive intervention performance above baseline then is not enough to determine that a model has low concept-residual information leakage. We postulate the following: for a model trained on a complete concept set, negative concept interventions should achieve near random performance. Models achieving higher accuracy by necessity must be using information in the residual layer. Looking at Figure \ref{fig:overall}, we see that for CUB, MI and IterNorm have similar intervention performance; however, their negative concept intervention accuracies are very different. While MI does not completely decorrelate the model, its performance of 19\% is still starkly lower then that of IterNorm, which achieves 60\% accuracy with random concepts. For incomplete concept sets like CIFAR 100, we see all methods but MI degrade substantially. This provides reasoning for why MI is able to plateau positive intervention in performance while other methods degrade. These findings indicate that positive interventions alone cannot fully determine the robustness and reliability of these models.

\begin{table}[h]
\centering
\specialsize
\caption{Correlation Coefficient Values ($r^2$) between Positive Intervention Accuracy and Disentanglement Metrics, Grouped by Residual Size. MI estimation uses CLUB, trained for 5 epochs posthoc for all methods \cite{cheng2020club}.}
\label{tab:pos_intervention_vs_disentanglement}
\begin{tabular}{c|cc|cc|cc}
\toprule
& \multicolumn{2}{c|}{CIFAR 100} & \multicolumn{2}{c|}{CUB} & \multicolumn{2}{c}{OAI} \\
\midrule
\makecell{Res.\\Size} & \makecell{$r^2$\\CC} & \makecell{$r^2$\\MI} & \makecell{$r^2$\\CC} & \makecell{$r^2$\\MI} & \makecell{$r^2$\\CC} & \makecell{$r^2$\\MI} \\
\midrule
1 & \textbf{0.191} & 0.103 & \textbf{0.276} & 0.218 & \textbf{0.533} & 0.433 \\
2 & 0.007 & \textbf{0.050} & \textbf{0.494} & 0.023 & 0.363 & \textbf{0.394} \\
4 & 0.109 & \textbf{0.564} & \textbf{0.701} & 0.081 & 0.053 & \textbf{0.440} \\
8 & 0.129 & \textbf{0.882} & \textbf{0.525} & 0.307 & 0.294 & \textbf{0.834} \\
16 & 0.033 & \textbf{0.863} & 0.298 & \textbf{0.569} & 0.214 & \textbf{0.541} \\
32 & 0.004 & \textbf{0.792} & \textbf{0.491} & 0.438 & 0.268 & \textbf{0.597} \\
64 & N/A & N/A & 0.436 & \textbf{0.602} & 0.137 & \textbf{0.551} \\
\bottomrule
\end{tabular}
\end{table}

\paragraph{Negative residual interventions.}
\textit{Negative residual interventions} offer additionally insights into the effectiveness of our decorrelation techniques, particularly when negative concept interventions are similar between two models. Ideally, negative residual intervention accuracies should approach baseline performance in datasets with a complete concept set, suggesting minimal information leakage. In Table \ref{fig:overall}, for OAI, MI Residual's performance nearly matches the baseline, indicating negligible reliance on the residual layer. Conversely, other methods, particularly IterNorm, perform poorly, indicating dependence on the residual layer. This explains IterNorm's inferior performance compared to MI Residual in positive interventions, despite similar outcomes in negative concept interventions.

\paragraph{Analyzing method vs. disentanglement metric correlation.}
\textit{Mutual information} (MI) and \textit{cross-correlation} (CC) are metrics for analyzing dependence between random variables \cite{song2012comparison}. Mahinpe et al. propose reducing mutual information to mitigate information leakage \cite{mahinpei2021promises}. We examine if MI and CC predict intervention performance (see Table \ref{tab:pos_intervention_vs_disentanglement}). A linear regression identifies the relationship between each metric and intervention performance. Our results show MI, except in small residuals, strongly correlates with performance, while CC tends to be a weaker predictor.

\section{Conclusion}
This paper offers a detailed study of CRBMs, emphasizing benchmarking and minimizing concept-residual leakage. We introduce \textit{semi-independent training}, a novel approach that restricts gradient backpropagation exclusively to the residual layer, effectively preventing intra-concept leakage and focusing on concept-residual leakage. Our three intervention methods—positive concept, random concept, and random residual—provide a thorough framework for assessing concept-residual leakage in CRBMs. Each method specifically examines different aspects of model functionality, facilitating a nuanced analysis of the model's ability to integrate and utilize concept and residual information.

For complete concept sets, the best CRBM decorrelation methods equaled baseline performance, signifying successful management of concept-residual leakage. Nevertheless, positive interventions revealed performance declines due to leakage. With incomplete concept sets, CRBMs surpassed bottleneck models in performance, particularly with performant concepts where Mutual Information (MI) significantly enhanced intervention outcomes. Conversely, nonperformant, incomplete sets displayed performance improvements as residual size increased, reflecting the limited influence of the concept layer. Negative interventions indicated that positive interventions alone are insufficient for detecting concept-residual leakage. Our findings also revealed that MI, particularly in larger residuals, outperformed cross-correlation in predicting leakage.

In summary, MI emerged as the most effective method in reducing concept-residual leakage, though it did not completely eliminate it. Future research can build on our methodology to develop more effective techniques for mitigating concept-residual leakage. Also, while we identified CIFAR-100 as an example of a performant, incomplete concept set, further work is necessary to benchmark CRBMs against a broader range of similar concept sets.

{\small
\bibliographystyle{ieee_fullname}
\bibliography{paper}
}

\end{document}